\documentclass[pdflatex,sn-mathphys-num]{sn-jnl}

\usepackage{graphicx}%
\usepackage{multirow}%
\usepackage{amsmath,amssymb,amsfonts}%
\usepackage{amsthm}%
\usepackage{mathrsfs}%
\usepackage[title]{appendix}%
\usepackage{xcolor}%
\usepackage{textcomp}%
\usepackage{manyfoot}%
\usepackage{booktabs}%
\usepackage{algorithm}%
\usepackage{algorithmicx}%
\usepackage{algpseudocode}%
\usepackage{listings}%
\usepackage{tikz} 
\usepackage{subfig}
\usepackage{colortbl}

\usepackage{array}
\usepackage{multicol}

\DeclareMathOperator*{\argmin}{arg\,min}

\begin{document}

\title[Article Title]{Random Gradient Masking as a Defensive Measure to Deep Leakage in Federated Learning}


\author[1]{\fnm{Joon} \sur{Kim}}\email{joonkim1@berkeley.edu}

\author*[2]{\fnm{Sejin} \sur{Park}}\email{baksejin@kmu.ac.kr}

\affil[1]{\orgdiv{Dept. of Electrical Engineering and Computer Science}, \orgname{UC Berkeley}, \orgaddress{\city{Berkeley}, \postcode{94720}, \state{CA}, \country{United States}}}

\affil*[2]{\orgdiv{Computer Science Dept.}, \orgname{Keimyung University}, \orgaddress{\street{2800, Dalgubeoldaero,}, \city{Dalseogu}, \state{Dalseogu}, \country{South Korea}}}


\abstract{Federated Learning(FL), in theory, preserves privacy of individual clients' data while producing quality machine learning models. However, attacks such as Deep Leakage from Gradients(DLG) severely question the practicality of FL. In this paper, we empirically evaluate the efficacy of four defensive methods against DLG: Masking, Clipping, Pruning, and Noising. Masking, while only previously studied as a way to compress information during parameter transfer, shows surprisingly robust defensive utility when compared to the other three established methods. Our experimentation is two-fold. We first evaluate the minimum hyperparameter threshold for each method across MNIST, CIFAR-10, and lfw datasets. Then, we train FL clients with each method and their minimum threshold values to investigate the trade-off between DLG defense and training performance. Results reveal that Masking and Clipping show near to none degradation in performance while obfuscating enough information to effectively defend against DLG.}




\maketitle

\section{Introduction}\label{sec1}

Federated Learning (FL)\citep{mcmahan_communication-efficient_2023}\citep{yang_federated_2019} emerged as an artificial intelligence training method that does not require sending data from peripheral devices(clients) to a central server. Rather, each client would download the central model from the server, train it over their private data, and send the resulting gradients of the private training back to the server, all of which are aggregated by a server-side algorithm to produce the next iteration of the central model. Ideally, mutually distrusted clients never communicate their private data, and yet they produce a central model that encompasses the entire clients’ data. Extensive research is being conducted on optimizing the learning efficiency of FL on various aspects such as incentive mechanisms\citep{liu_gtg-shapley_2021}, communication speed\citep{sun_communication-efficient_2019}, non-IID training\citep{li_federated_2021}, and client selection\citep{cho_towards_2022}.

However, recent research reveals that sending the gradients of private training does not ensure complete data privacy, especially in a wide cross-device environment\citep{shejwalkar_back_2021}. Moreover, as a federated system, FL has to protect itself against Byzantine Failure\citep{blanchard_machine_2017}, Backdoor injection\citep{bagdasaryan_how_2019}, Model Poisoning\citep{bhagoji_analyzing_2019}, and Data Poisoning\citep{tolpegin_data_2020}). In this paper, we propose random gradient masking as a defensive measure against DLG. While other studies have introduced random masking\cite{konecny_federated_2017}, they have primarily mentioned it as a defense to BFT or an improvement to communication efficiency. We aim to reexamine the efficacy of masking for protecting privacy under data leakage in comparison to other state-of-the-art methods such as noising, clipping, and pruning. 

We will first evaluate defensive thresholds for each method against iDLG and evaluate convergence in i.i.d. and non-i.i.d. settings to compare the security-performance trade-off between methods. Defensive thresholds were measured empirically on CIFAR-10, MNIST, and lfw images with ResNet-18 using the SSIM metric between original and leaked images.
Convergence tests were performed using the CIFAR-10 dataset on ResNet-20.

Technical contributions of this paper include:
\begin{itemize}
    \item Proposing Random Masking as a technique to defend against DLG
    \item Evaluating defensive performance against iDLG with SSIM of Masking, Clipping, Pruning, and Noising to empirically measure threshold hyperparameters
    \item Comparing convergence of Masking, Noising, Pruning, and Clipping based on the threshold measured in the previous experiment
\end{itemize}

\section{Background}\label{sec:background}
\subsection{The Naive FL Model}

\begin{algorithm}[H]
    \caption{Naive Federated Learning}
    \label{alg:FL}
    \begin{algorithmic}
        \Require Central Server $S$, Central Model $M$, Client Models $C_1...C_n$, Client Data $D_1...D_n$, Aggregation Function $Agg(\cdot)$, Number of Rounds $R$, Epochs per Round $E$
        \Ensure Each Client Data $D_i$ is only available to client $C_i$

        \State $M \gets$ random initialization by $S$
        \For{$r$ in $1\ldots R$}
            \For{$C_i$ in $C_1\ldots C_n$ in parallel} 
                \State $C_i\gets M$ 
                \For{$e$ in $1\ldots E$}
                    \State $Fit(C_i, D_i)$ 
                \EndFor
                \State Send $Parameters(C_i)$ to $S$
                \EndFor
            \State $M\gets Agg(C_1,\ldots,C_n)$ 
            \EndFor
            \State \Return $M$
        
    \end{algorithmic}
\end{algorithm}

In an FL framework, all clients possess private data not shared to any other client or the central server. At the start of each round, the client receives parameters for the full model, which is identical for all clients. The clients proceed to train the model utilizing their private data, resulting in variation of models amongst clients. Once the training is finished, each client submits their trained parameters back to the server, which then aggregates the submissions through an aggregation algorithm. Common algorithms include (Multi-)KRUM\citep{yin_byzantine-robust_2021}, FedAvg(Weighted Average)\citep{blanchard_machine_2017}. The aggregated parameters are then disseminated to all clients for local evaluation and training of the next round. The expectation is that even though the training data are disjoint amongst clients, a well-defined set of training instructions and aggregation could lead to convergence of the central model.

\subsection{Deep Leakage from Gradients}

Deep Leakage from Gradients(DLG)\citep{zhu_deep_2019} aims to reverse-engineer private training data based on client-submitted gradients. Before the findings of DLG, it was believed that the gradient values alone did not possess enough information to replicate the training data. However, subsequent research\citep{zhao_idlg_2020}\citep{geiping_inverting_2020}\citep{yin_see_2021} revealed that DLG techniques can be developed to leak commonly adopted defensive algorithms. With such techniques, third-party eavesdroppers or even the central server itself can recreate the clients’ private data, which defeats the purpose of FL, which is to keep private data out of the third party's reach. The general flow of DLG is as follows:

\begin{multline}
    x'^*, y'^* = \argmin_{x',y'}{||\nabla W'- G||^2} = \argmin_{x',y'}{||\frac{\partial L(F(x',W),y')}{\partial W}- G||^2}
\end{multline}

To defend against DLG, clients inevitably have to manipulate their submission before transmitting it to the server, incurring a fundamental trade-off between privacy and training performance. For example, Differential Privacy(DP)\citep{wei_federated_2019}\citep{agarwal_skellam_2021} adds random Gaussian noise to the training result to disrupt the reversing process. However, the variance of noise needed to successfully prevent DLG causes a significant slowdown of the overall training. Gradient Compression(Pruning)\citep{mondal_beas_2022} removes gradient values smaller than a threshold value, and Gradient Clipping\citep{panda_sparsefed_2021} sets all values bigger than the threshold to the threshold value. While Pruning and Clipping are efficient methods to prevent DLG, they require experimentation on each specific model and dataset setting to determine the appropriate threshold value, a hyperparameter. Other attempted methods include homomorphic encryption(HE)\citep{zhang_batchcrypt_nodate} and trusted execution environment(TEE)\citep{mo_ppfl_2021}.

\section{Methodology}
\subsection{Client-Side Obfuscation}

The most intuitive and effective defense against DLG is to purposefully obfuscate the client submissions before transmitting. The requirement is that the loss of information is irreversible with only one client submission but the server can recover the same distribution of parameters when aggregating with multiple submissions. This study examines the defensive performance of four algorithms, three already established as effective defenses (Noising, Clipping, Pruning) and one newly proposed (Masking). The full pseudocode of each method can be found in Appendix A.

\begin{itemize}
    \item \textbf{Noising} adds a random value to each parameter extracted from a predefined distribution, usually Gaussian.
    \item \textbf{Clipping} determines a maximum threshold value for each layer and sets all parameters with greater absolute value than the threshold to the threshold value.
    \item \textbf{Pruning}, also referred to as compression, determines a minimum threshold value for each layer and sets all parameters with less absolute values than the threshold to 0.
    \item \textbf{Masking} randomly sets a certain ratio of values to NaN such that it would not be included in the process of aggregation.
\end{itemize}

\subsection{BatchNorm in Masking}

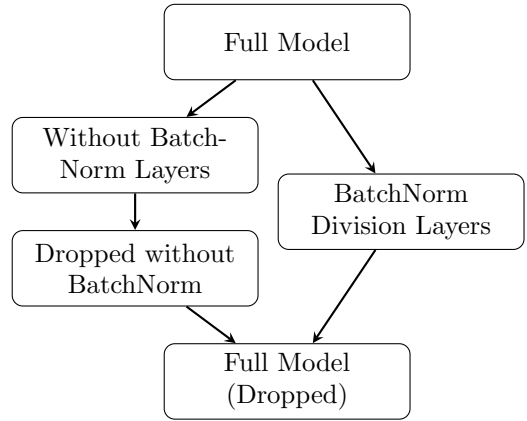
\begin{figure}[!htb]
\centering
\tikzstyle{block} = [rectangle, rounded corners, minimum width=3cm, minimum height=1cm,text centered, draw=black, text width=3cm]
\begin{tikzpicture}[node distance=1.5cm]
\node (start) [block] {Full Model};
\node (woBatch) [block, below of=start, xshift=-2cm] {Without BatchNorm Layers};
\node (batch) [block, right of=woBatch, xshift=2cm, yshift=-0.75cm] {BatchNorm Division Layers};
\node (dropped) [block, below of=woBatch] {Dropped without BatchNorm};
\node (end) [block, below of=dropped, xshift=2cm] {Full Model (Dropped)};

\tikzstyle{arrow} = [thick,->,>=stealth]
\draw [arrow] (start) -- (woBatch);
\draw [arrow] (start) -- (batch);
\draw [arrow] (woBatch) -- (dropped);
\draw [arrow] (dropped) -- (end);
\draw [arrow] (batch) -- (end);
\end{tikzpicture}
\caption{Flowchart Diagram for BatchNorm Circumvention in Masking}
\vspace{-0.5cm}
\label{fig:batchnorm_diagram}
\end{figure}

An obscure problem of masking random values is that certain layers might malfunction from the absence of values. One such is included in the widely used BatchNorm\citep{ioffe_batch_2015}: the parameters used as denominators and multipliers for normalizing. During experimentation, we discovered that masking does not converge at all for models including BatchNorm layers. At first, experiments were carried out by removing all BatchNorm layers. However, removing layers caused a significant decrease in convergence. The eventual solution was to separate the layers for BatchNorm division layers and only apply masking to the remaining layers. Then, the BatchNorm division layers are recombined with the rest to produce the full dropped model. Figure \ref{fig:batchnorm_diagram} illustrates the technique outlined above. This approach reduces the actual percentage of the masked values but is the most general solution to incompatible layers. All such layers that cannot afford any of its values removed should also adopt this procedure. The circumvention does not affect the performance of the masking since ResNet architectures allocate only around 1\% of the total parameters for batch normalization.

\section{Experiment Design}

\subsection{Threshold Measurement}\label{sec:threshold_measurement_experiments}

\begin{figure}[h!]%
    \centering
    \subfloat[\centering Masking 20\%]{{\includegraphics[width=0.9\linewidth]{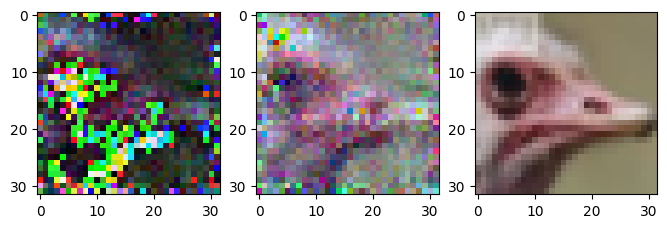} }}%

    \subfloat[\centering Pruning 80\%]{{\includegraphics[width=0.9\linewidth]{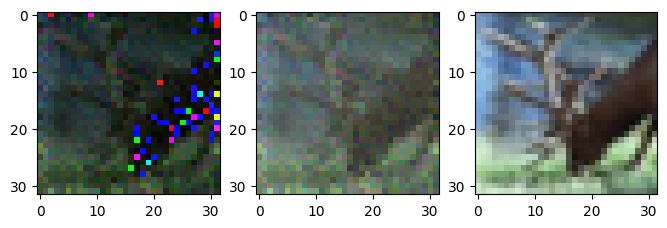} }}%
    
    \caption{iDLG that are leaked with brightness adjustments. The leftmost image is the leaked image, and the middle image is the one scoring best in SSIM compared to the original image, out of all brightness-adjusted images. The leaked image at first seems obfuscated enough, but with adjustments, it is leaked.}%
    \label{fig:leakage_with_adjustments_leaked}%
\end{figure}

\begin{figure}[h!]%
    \centering
    \subfloat[\centering Masking 40\%]{{\includegraphics[width=0.9\linewidth]{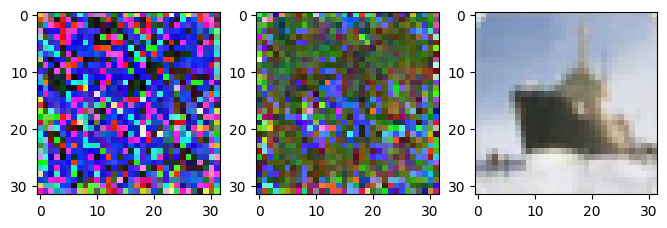} }}%

    \subfloat[\centering Pruning 95\%]{{\includegraphics[width=0.9\linewidth]{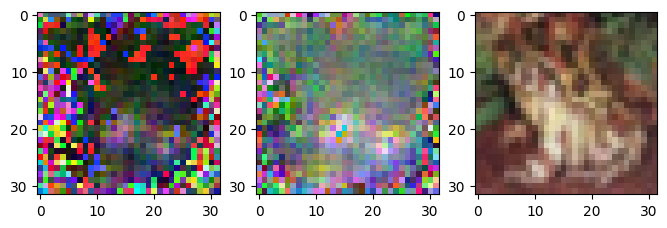} }}%
    
    \caption{iDLG that are robust even with brightness adjustments. The obfuscation methods are identical to Figure \ref{fig:leakage_with_adjustments_leaked}, but with hyperparameters intensified to obfuscate more information against the third party.}%
    \label{fig:leakage_with_adjustments_defended}%
\end{figure}

Three datasets, MNIST, CIFAR-10, and lfw, alongside ResNet-18, were used in evaluating the threshold for each algorithm’s unique hyperparameter. MNIST and CIFAR-10 images were used as 28x28 and 32x32 pixel images as intended, but lfw images were compressed into 32x32 images by the torchvision.transforms.Resize function. The leakage algorithm of interest was iDLG\cite{zhao_idlg_2020}, an improvement from the original DLG algorithm. Unlike the original iDLG algorithm, we elected Adam as the optimizer, with a learning rate of 0.03 and decay of 0.01. The evaluation performs the iDLG method until the mean squared error does not decrease for two consecutive 30-round checkpoints. 

During the experiment, we realized that comparing the raw output of iDLG with the original image does not accurately reflect the degree of leakage when analyzed by the human eye. This was mainly due to the tendency of iDLG to decrease the pixel values throughout its leakage, making the entire image darker and, for some pixels, wrap around the 0-255 range to become extremely high values. Such phenomenon is well shown in Figure \ref{fig:leakage_with_adjustments_leaked}, where some dark portions of the image are represented in the iDLG output as bright pixels instead. A well-performing defense should behave similar to examples in Figure \ref{fig:leakage_with_adjustments_defended} instead.

Thus, we decided to evaluate the actual degree of leakage by adjusting the brightness of the resulting image. This adjustment involves adding values ranging from 10 to 200 to the resulting image, creating 20 augmented images. The Structural Similarity Index Measure (SSIM) is then used to compare the augmented images to the original to find the maximum SSIM score. The SSIM score assesses the visual similarity between two images, which helps determine the degree to which the leaked image, after brightness adjustment, resembles the original. The hardware used for threshold measurement was primarily a T4 GPU provided by Google Colaboratory.

\subsection{Performance Evaluation}

The convergence evaluation aims to compare the convergence properties of various methods under the same training environment. Specifically, we seek to understand how different methods perform throughout reaching a stable accuracy over a given number of rounds. The objective is to determine which method converges faster or more reliably under identical training conditions and data distributions.

The evaluation uses the CIFAR-10 dataset with ResNet-20. The data distribution includes both Independent and Identically Distributed (IID) and Non-Independent and Identically Distributed (non-iid) with beta=0.5. In the IID case, each client receives a balanced subset of the data. In the non-iid case, the data distribution among clients is skewed, meaning each client may have a biased subset of the data, controlled by the Dirichlet distribution parameter beta. All evaluation involves 10 clients participating in the training process over 25 training rounds.

All methods are trained using the Adam optimizer with a learning rate of 0.001. All methods covered in the threshold evaluation are compared, which might include various federated learning algorithms, gradient descent variations, or other optimization techniques used in distributed settings. The experiment is conducted within the same training environment to ensure consistency. The hardware used for performance evaluation was primarily RTX3090 and RTX A6000 provided by RunPod.io.

\section{Discussion}

\subsection{Threshold Measurement}

\begin{table}[ht!]
\centering
\small
\begin{tabular}{|c|c|c|c|c|}
\hline
          & Metrics & MNIST & CIFAR & lfw \\ \hline
Original & N/A            & 0.75          & 1.0             & 0.98         \\ \hline
  & 0.05          & 0.77          & 0.95            & 0.98         \\ 
Noising                  & 0.25          & 0.5           & 0.75            & 0.64         \\ 
                  & \cellcolor[HTML]{FFDD83}0.5 & \cellcolor[HTML]{FFDD83}0.38 & \cellcolor[HTML]{FFDD83}0.50 & \cellcolor[HTML]{FFDD83}0.27 \\ \hline
 & 0.999         & 0.10          & 0.84            & 0.84         \\  
Clipping                  & \cellcolor[HTML]{FFDD83}0.995 & \cellcolor[HTML]{FFDD83}0.10 & \cellcolor[HTML]{FFDD83}0.12 & \cellcolor[HTML]{FFDD83}0.09 \\  
                  & 0.990         & 0.04          & 0.09            & 0.09         \\ \hline
  & 0.8           & 0.52          & 0.72            & 0.73         \\ 
Pruning                  & 0.9           & 0.34          & 0.53            & 0.65         \\  
                  & \cellcolor[HTML]{FFDD83}0.95 & \cellcolor[HTML]{FFDD83}0.18 & \cellcolor[HTML]{FFDD83}0.28 & \cellcolor[HTML]{FFDD83}0.36 \\ \hline
  & 0.2           & 0.33          & 0.64            & 0.34         \\  
Masking                  & 0.3           & 0.13          & 0.39            & 0.44         \\  
                  & \cellcolor[HTML]{FFDD83}0.4  & \cellcolor[HTML]{FFDD83}0.10 & \cellcolor[HTML]{FFDD83}0.13 & \cellcolor[HTML]{FFDD83}0.16 \\ \hline
\end{tabular}
\caption{Comparison of threshold values across different datasets.}
\label{threshold-table}
\end{table}

\begin{figure*}[ht!]%
    \centering
    
    \centering
    \subfloat[\centering CIFAR-10]{{\includegraphics[width=0.9\linewidth]{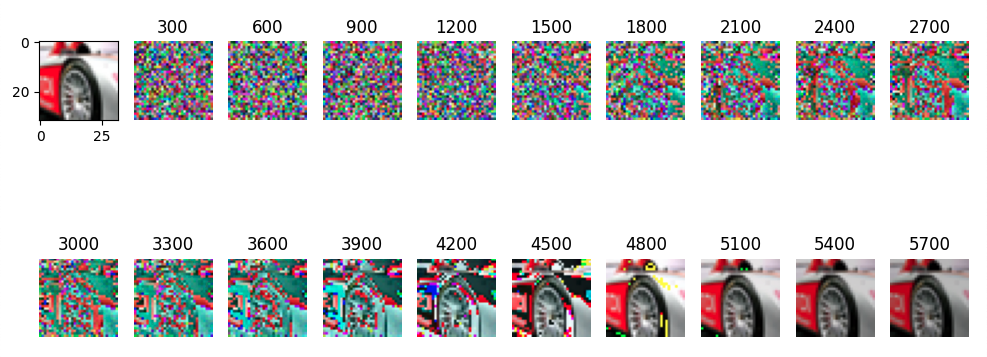} }}\\
    \subfloat[\centering MNIST]{{\includegraphics[width=0.45\linewidth]{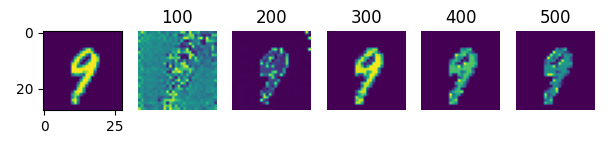} }}%
    \qquad
    \centering
    \subfloat[\centering lfw]{{\includegraphics[width=0.45\linewidth]{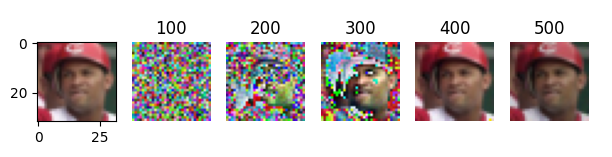} }}%
    
    \caption{iDLG without any obfuscation}%
    \label{fig:raw_leakage}%
\end{figure*}

Table \ref{threshold-table} shows the result of evaluating several different values of hyperparameters for each obfuscation algorithm across three distinct datasets. Clipping and Masking work well with hyperparameters of 0.995 and 0.4, respectively. However, Noising and Pruning algorithms do not show a drastic cut-off after a certain threshold; experiments were stopped after hyperparameters of 0.5 and 0.95, respectively. As presented in the next section, values lower than the maximum values tested, 0.25 and 0.9, are enough to impact the convergence of FL training significantly. There was no need to continue experimenting with higher hyperparameter values for Noising and Pruning, as it only exacerbated the training performance.

Analyzing the results column-wise, MNIST, which has only one channel instead of three RGB channels, shows higher resistance to iDLG. Such a result is surprising since MNIST originally possesses less information required to be leaked by the iDLG algorithm and would be expected to be more prone to leakage. However, examining the actual leaked images gives insight into its reported robustness.

Figure \ref{fig:raw_leakage} shows iDLG attacks on each dataset without any obfuscation methods applied. As mentioned in Section \ref{sec:threshold_measurement_experiments}, there is a "graying" effect of iDLG on the leaked images, which are compensated by additions in pixel values. For MNIST specifically, however, the background pixels stay black without underflowing into white pixels. Effectively, only the handwritten digits are affected by the "graying," which leads to lower SSIM scores. Nonetheless, MNIST images represent a realistic iDLG attack scenario, and the SSIM scores calculated are consistent with the human eye interpretation of the degree of leakage. As a side note, lfw images tended to be leaked faster than CIFAR-10 images, just as shown in Figures \ref{fig:raw_leakage}(a) and \ref{fig:raw_leakage}(c). We conjecture that as lfw images are compressed into 32x32 images, they tend to contain less information needed to reconstruct the image than the originally 32x32 rendered CIFAR-10 images.

\begin{figure*}[h!]%
    \centering
    \subfloat[\centering IID Convergence]{{\includegraphics[width=0.9\linewidth]{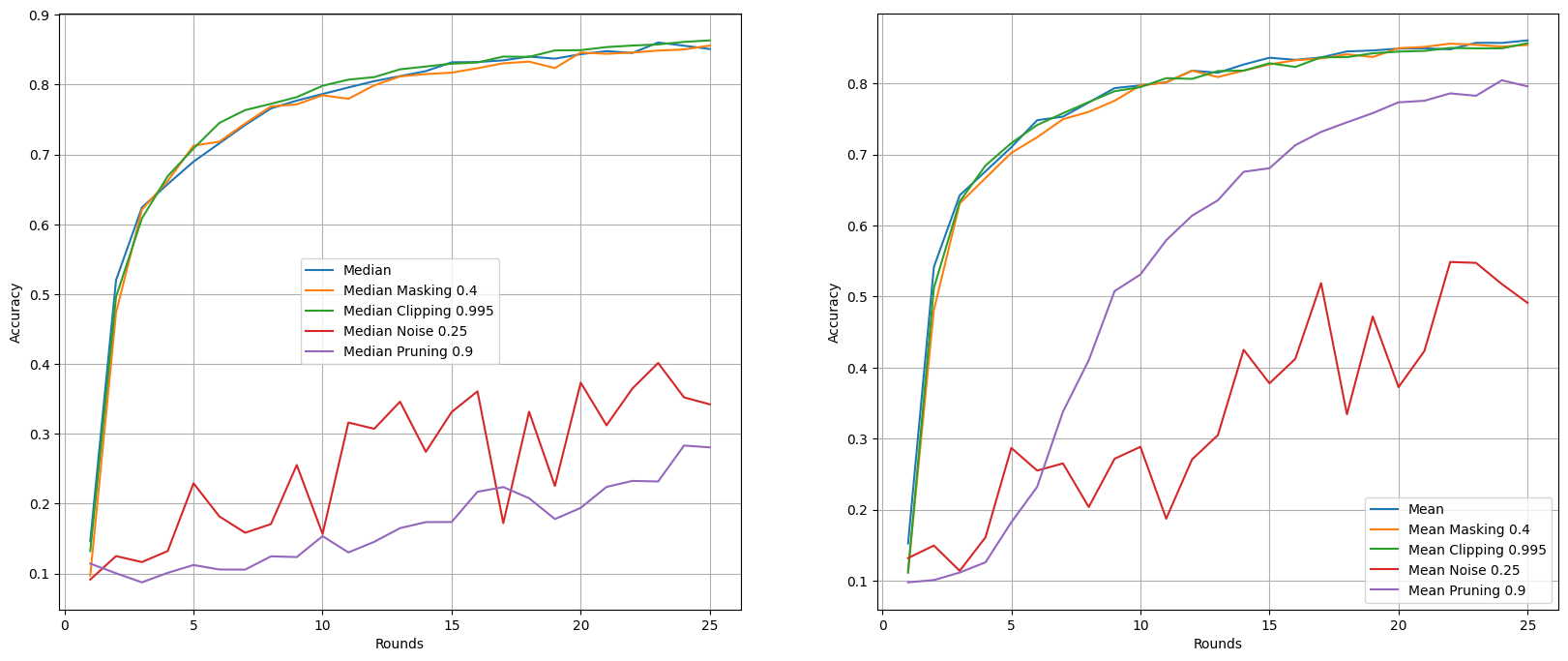} }}%

    \subfloat[\centering Non-IID Convergence ]{{\includegraphics[width=0.9\linewidth]{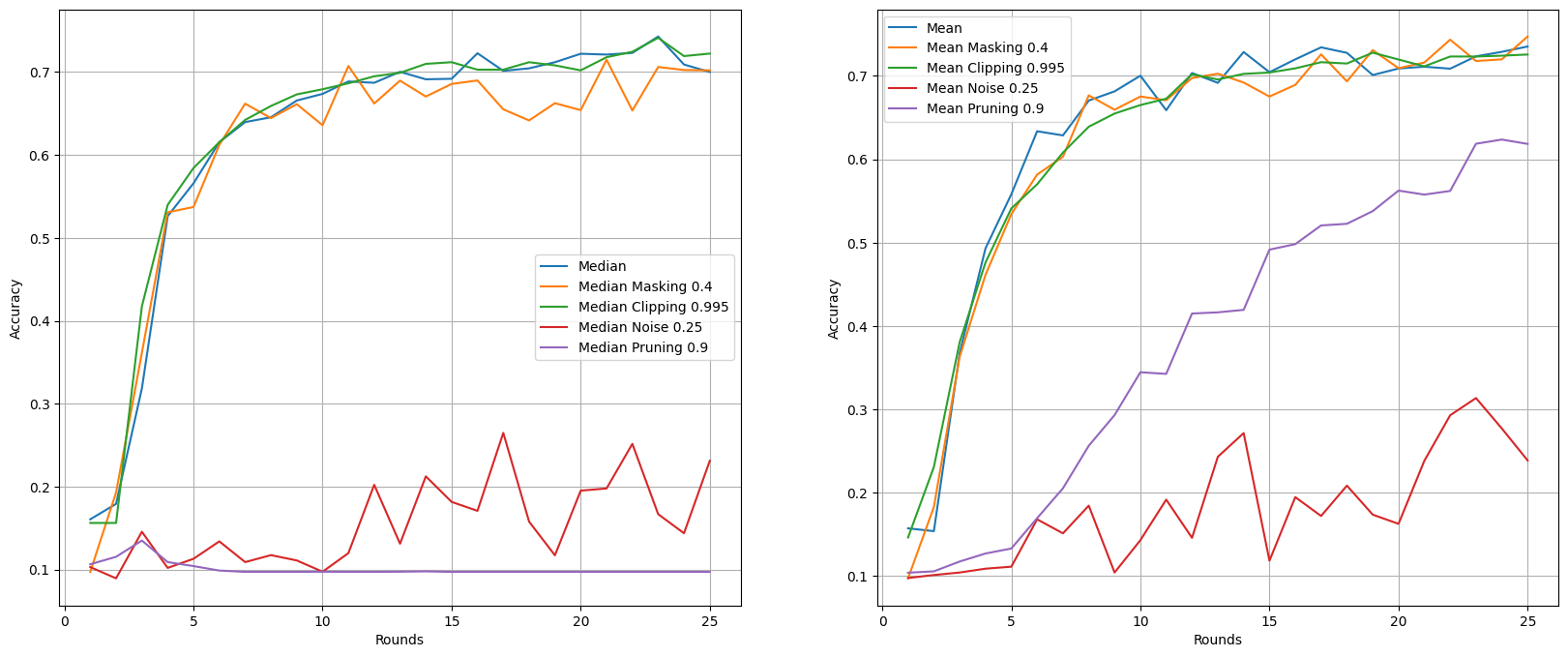} }}%
    
    \caption{\centering Performance Evaluation of various obfuscation algorithms with CIFAR-10 for Median(Left) and Mean(Right), IID(Up) and Non-IID(Down).}%
    \label{fig:performance_evaluation}%
\end{figure*}

\subsection{Performance Evaluation}

Figure \ref{fig:performance_evaluation} illustrates the evaluation accuracy of each obfuscation method compared to non-obfuscated FL. Four graphs each represent a combination of the Median or Mean aggregation method with IID or Non-IID settings of the dataset. Due to the nature of unstable Non-IID datasets, they have lower final converging accuracy and are more volatile when measured each round. Nonetheless, it is recognizable that Masking(0.4) and Clipping(0.995) perform well across all possible combinations and matches the accuracy curve of the non-obfuscated training closely. Pruning and Noising, on the other hand, even with hyperparameters resulting in subpar defensive utility, significantly degrade the training performance. It is worth noting that Pruning performs worse in Median aggregation than Noising but outperforms it in Mean aggregation settings, reaching $\sim$10\% difference compared to other well-performing methods. 

\section{Conclusion}

The Deep Leakage from Gradients attack undermines the motivation behind FL, and it is critical to prevent it through obfuscating the client submissions before communicating with the central server. The challenge is that the trade-off between robustness in security and training performance is inevitable. Through experimentation, we were able to quantitatively measure and compare the trade-offs of commonly used methods such as Clipping, Pruning, and Noising, as well as Masking, a technique previously only applied in information compression. Our results reveal that Masking and Clipping are ideal forms of obfuscation, as they minimize the damage to training performance while maintaining robustness in defending against DLG attacks.

\bibliography{main}


\begin{thebibliography}{24}
\ifx \bisbn   \undefined \def \bisbn  #1{ISBN #1}\fi
\ifx \binits  \undefined \def \binits#1{#1}\fi
\ifx \bauthor  \undefined \def \bauthor#1{#1}\fi
\ifx \batitle  \undefined \def \batitle#1{#1}\fi
\ifx \bjtitle  \undefined \def \bjtitle#1{#1}\fi
\ifx \bvolume  \undefined \def \bvolume#1{\textbf{#1}}\fi
\ifx \byear  \undefined \def \byear#1{#1}\fi
\ifx \bissue  \undefined \def \bissue#1{#1}\fi
\ifx \bfpage  \undefined \def \bfpage#1{#1}\fi
\ifx \blpage  \undefined \def \blpage #1{#1}\fi
\ifx \burl  \undefined \def \burl#1{\textsf{#1}}\fi
\ifx \doiurl  \undefined \def \doiurl#1{\url{https://doi.org/#1}}\fi
\ifx \betal  \undefined \def \betal{\textit{et al.}}\fi
\ifx \binstitute  \undefined \def \binstitute#1{#1}\fi
\ifx \binstitutionaled  \undefined \def \binstitutionaled#1{#1}\fi
\ifx \bctitle  \undefined \def \bctitle#1{#1}\fi
\ifx \beditor  \undefined \def \beditor#1{#1}\fi
\ifx \bpublisher  \undefined \def \bpublisher#1{#1}\fi
\ifx \bbtitle  \undefined \def \bbtitle#1{#1}\fi
\ifx \bedition  \undefined \def \bedition#1{#1}\fi
\ifx \bseriesno  \undefined \def \bseriesno#1{#1}\fi
\ifx \blocation  \undefined \def \blocation#1{#1}\fi
\ifx \bsertitle  \undefined \def \bsertitle#1{#1}\fi
\ifx \bsnm \undefined \def \bsnm#1{#1}\fi
\ifx \bsuffix \undefined \def \bsuffix#1{#1}\fi
\ifx \bparticle \undefined \def \bparticle#1{#1}\fi
\ifx \barticle \undefined \def \barticle#1{#1}\fi
\bibcommenthead
\ifx \bconfdate \undefined \def \bconfdate #1{#1}\fi
\ifx \botherref \undefined \def \botherref #1{#1}\fi
\ifx \url \undefined \def \url#1{\textsf{#1}}\fi
\ifx \bchapter \undefined \def \bchapter#1{#1}\fi
\ifx \bbook \undefined \def \bbook#1{#1}\fi
\ifx \bcomment \undefined \def \bcomment#1{#1}\fi
\ifx \oauthor \undefined \def \oauthor#1{#1}\fi
\ifx \citeauthoryear \undefined \def \citeauthoryear#1{#1}\fi
\ifx \endbibitem  \undefined \def \endbibitem {}\fi
\ifx \bconflocation  \undefined \def \bconflocation#1{#1}\fi
\ifx \arxivurl  \undefined \def \arxivurl#1{\textsf{#1}}\fi
\csname PreBibitemsHook\endcsname

\bibitem[\protect\citeauthoryear{{McMahan} et~al.}{}]{mcmahan_communication-efficient_2023}
\begin{botherref}
\oauthor{\bsnm{{McMahan}}, \binits{H.B.}},
\oauthor{\bsnm{Moore}, \binits{E.}},
\oauthor{\bsnm{Ramage}, \binits{D.}},
\oauthor{\bsnm{Hampson}, \binits{S.}},
\oauthor{\bsnm{Arcas}, \binits{B.A.y.}}:
Communication-Efficient Learning of Deep Networks from Decentralized Data.
{arXiv}.
\doiurl{10.48550/arXiv.1602.05629} .
\url{http://arxiv.org/abs/1602.05629}
Accessed 2024-04-17
\end{botherref}
\endbibitem

\bibitem[\protect\citeauthoryear{Yang et~al.}{}]{yang_federated_2019}
\begin{botherref}
\oauthor{\bsnm{Yang}, \binits{Q.}},
\oauthor{\bsnm{Liu}, \binits{Y.}},
\oauthor{\bsnm{Chen}, \binits{T.}},
\oauthor{\bsnm{Tong}, \binits{Y.}}:
Federated machine learning: Concept and applications
\textbf{10}(2),
12--11219
\doiurl{10.1145/3298981} .
Accessed 2024-04-04
\end{botherref}
\endbibitem

\bibitem[\protect\citeauthoryear{Liu et~al.}{}]{liu_gtg-shapley_2021}
\begin{botherref}
\oauthor{\bsnm{Liu}, \binits{Z.}},
\oauthor{\bsnm{Chen}, \binits{Y.}},
\oauthor{\bsnm{Yu}, \binits{H.}},
\oauthor{\bsnm{Liu}, \binits{Y.}},
\oauthor{\bsnm{Cui}, \binits{L.}}:
{GTG}-Shapley: Efficient and Accurate Participant Contribution Evaluation in Federated Learning.
{arXiv}.
\url{http://arxiv.org/abs/2109.02053}
Accessed 2023-12-10
\end{botherref}
\endbibitem

\bibitem[\protect\citeauthoryear{Sun et~al.}{}]{sun_communication-efficient_2019}
\begin{botherref}
\oauthor{\bsnm{Sun}, \binits{J.}},
\oauthor{\bsnm{Chen}, \binits{T.}},
\oauthor{\bsnm{Giannakis}, \binits{G.B.}},
\oauthor{\bsnm{Yang}, \binits{Z.}}:
Communication-Efficient Distributed Learning via Lazily Aggregated Quantized Gradients.
{arXiv}.
\url{http://arxiv.org/abs/1909.07588}
Accessed 2023-12-10
\end{botherref}
\endbibitem

\bibitem[\protect\citeauthoryear{Li et~al.}{}]{li_federated_2021}
\begin{botherref}
\oauthor{\bsnm{Li}, \binits{Q.}},
\oauthor{\bsnm{Diao}, \binits{Y.}},
\oauthor{\bsnm{Chen}, \binits{Q.}},
\oauthor{\bsnm{He}, \binits{B.}}:
Federated Learning on Non-{IID} Data Silos: An Experimental Study.
{arXiv}.
\url{http://arxiv.org/abs/2102.02079}
Accessed 2023-11-26
\end{botherref}
\endbibitem

\bibitem[\protect\citeauthoryear{Cho et~al.}{}]{cho_towards_2022}
\begin{botherref}
\oauthor{\bsnm{Cho}, \binits{Y.J.}},
\oauthor{\bsnm{Wang}, \binits{J.}},
\oauthor{\bsnm{Joshi}, \binits{G.}}:
Towards understanding biased client selection in federated learning.
In: Proceedings of The 25th International Conference on Artificial Intelligence and Statistics,
pp. 10351--10375.
{PMLR}.
{ISSN}: 2640-3498.
\url{https://proceedings.mlr.press/v151/jee-cho22a.html}
Accessed 2023-12-10
\end{botherref}
\endbibitem

\bibitem[\protect\citeauthoryear{Shejwalkar et~al.}{}]{shejwalkar_back_2021}
\begin{botherref}
\oauthor{\bsnm{Shejwalkar}, \binits{V.}},
\oauthor{\bsnm{Houmansadr}, \binits{A.}},
\oauthor{\bsnm{Kairouz}, \binits{P.}},
\oauthor{\bsnm{Ramage}, \binits{D.}}:
Back to the Drawing Board: A Critical Evaluation of Poisoning Attacks on Production Federated Learning.
{arXiv}.
\url{http://arxiv.org/abs/2108.10241}
Accessed 2023-11-30
\end{botherref}
\endbibitem

\bibitem[\protect\citeauthoryear{Blanchard et~al.}{}]{blanchard_machine_2017}
\begin{botherref}
\oauthor{\bsnm{Blanchard}, \binits{P.}},
\oauthor{\bsnm{El~Mhamdi}, \binits{E.M.}},
\oauthor{\bsnm{Guerraoui}, \binits{R.}},
\oauthor{\bsnm{Stainer}, \binits{J.}}:
Machine learning with adversaries: Byzantine tolerant gradient descent.
In: Advances in Neural Information Processing Systems,
vol. 30.
Curran Associates, Inc.
\url{https://papers.nips.cc/paper_files/paper/2017/hash/f4b9ec30ad9f68f89b29639786cb62ef-Abstract.html}
Accessed 2024-04-12
\end{botherref}
\endbibitem

\bibitem[\protect\citeauthoryear{Bagdasaryan et~al.}{}]{bagdasaryan_how_2019}
\begin{botherref}
\oauthor{\bsnm{Bagdasaryan}, \binits{E.}},
\oauthor{\bsnm{Veit}, \binits{A.}},
\oauthor{\bsnm{Hua}, \binits{Y.}},
\oauthor{\bsnm{Estrin}, \binits{D.}},
\oauthor{\bsnm{Shmatikov}, \binits{V.}}:
How To Backdoor Federated Learning.
\url{http://arxiv.org/abs/1807.00459}
Accessed 2023-12-07
\end{botherref}
\endbibitem

\bibitem[\protect\citeauthoryear{Bhagoji et~al.}{}]{bhagoji_analyzing_2019}
\begin{botherref}
\oauthor{\bsnm{Bhagoji}, \binits{A.N.}},
\oauthor{\bsnm{Chakraborty}, \binits{S.}},
\oauthor{\bsnm{Mittal}, \binits{P.}},
\oauthor{\bsnm{Calo}, \binits{S.}}:
Analyzing Federated Learning through an Adversarial Lens.
{arXiv}.
\url{http://arxiv.org/abs/1811.12470}
Accessed 2023-12-10
\end{botherref}
\endbibitem

\bibitem[\protect\citeauthoryear{Tolpegin et~al.}{}]{tolpegin_data_2020}
\begin{botherref}
\oauthor{\bsnm{Tolpegin}, \binits{V.}},
\oauthor{\bsnm{Truex}, \binits{S.}},
\oauthor{\bsnm{Gursoy}, \binits{M.E.}},
\oauthor{\bsnm{Liu}, \binits{L.}}:
Data Poisoning Attacks Against Federated Learning Systems.
{arXiv}.
\url{http://arxiv.org/abs/2007.08432}
Accessed 2023-12-10
\end{botherref}
\endbibitem

\bibitem[\protect\citeauthoryear{Konečný et~al.}{}]{konecny_federated_2017}
\begin{botherref}
\oauthor{\bsnm{Konečný}, \binits{J.}},
\oauthor{\bsnm{{McMahan}}, \binits{H.B.}},
\oauthor{\bsnm{Yu}, \binits{F.X.}},
\oauthor{\bsnm{Richtárik}, \binits{P.}},
\oauthor{\bsnm{Suresh}, \binits{A.T.}},
\oauthor{\bsnm{Bacon}, \binits{D.}}:
Federated Learning: Strategies for Improving Communication Efficiency.
{arXiv}.
\url{http://arxiv.org/abs/1610.05492}
Accessed 2024-04-26
\end{botherref}
\endbibitem

\bibitem[\protect\citeauthoryear{Yin et~al.}{}]{yin_byzantine-robust_2021}
\begin{botherref}
\oauthor{\bsnm{Yin}, \binits{D.}},
\oauthor{\bsnm{Chen}, \binits{Y.}},
\oauthor{\bsnm{Ramchandran}, \binits{K.}},
\oauthor{\bsnm{Bartlett}, \binits{P.}}:
Byzantine-Robust Distributed Learning: Towards Optimal Statistical Rates.
{arXiv}.
\url{http://arxiv.org/abs/1803.01498}
Accessed 2023-11-26
\end{botherref}
\endbibitem

\bibitem[\protect\citeauthoryear{Zhu et~al.}{}]{zhu_deep_2019}
\begin{botherref}
\oauthor{\bsnm{Zhu}, \binits{L.}},
\oauthor{\bsnm{Liu}, \binits{Z.}},
\oauthor{\bsnm{Han}, \binits{S.}}:
Deep Leakage from Gradients.
{arXiv}.
\url{http://arxiv.org/abs/1906.08935}
Accessed 2023-11-26
\end{botherref}
\endbibitem

\bibitem[\protect\citeauthoryear{Zhao et~al.}{}]{zhao_idlg_2020}
\begin{botherref}
\oauthor{\bsnm{Zhao}, \binits{B.}},
\oauthor{\bsnm{Mopuri}, \binits{K.R.}},
\oauthor{\bsnm{Bilen}, \binits{H.}}:
{iDLG}: Improved Deep Leakage from Gradients.
{arXiv}.
\url{http://arxiv.org/abs/2001.02610}
Accessed 2023-11-26
\end{botherref}
\endbibitem

\bibitem[\protect\citeauthoryear{Geiping et~al.}{}]{geiping_inverting_2020}
\begin{botherref}
\oauthor{\bsnm{Geiping}, \binits{J.}},
\oauthor{\bsnm{Bauermeister}, \binits{H.}},
\oauthor{\bsnm{Dröge}, \binits{H.}},
\oauthor{\bsnm{Moeller}, \binits{M.}}:
Inverting Gradients -- How easy is it to break privacy in federated learning?
{arXiv}.
\url{http://arxiv.org/abs/2003.14053}
Accessed 2023-11-26
\end{botherref}
\endbibitem

\bibitem[\protect\citeauthoryear{Yin et~al.}{}]{yin_see_2021}
\begin{botherref}
\oauthor{\bsnm{Yin}, \binits{H.}},
\oauthor{\bsnm{Mallya}, \binits{A.}},
\oauthor{\bsnm{Vahdat}, \binits{A.}},
\oauthor{\bsnm{Alvarez}, \binits{J.M.}},
\oauthor{\bsnm{Kautz}, \binits{J.}},
\oauthor{\bsnm{Molchanov}, \binits{P.}}:
See through Gradients: Image Batch Recovery via {GradInversion}.
{arXiv}.
\url{http://arxiv.org/abs/2104.07586}
Accessed 2023-11-26
\end{botherref}
\endbibitem

\bibitem[\protect\citeauthoryear{Wei et~al.}{}]{wei_federated_2019}
\begin{botherref}
\oauthor{\bsnm{Wei}, \binits{K.}},
\oauthor{\bsnm{Li}, \binits{J.}},
\oauthor{\bsnm{Ding}, \binits{M.}},
\oauthor{\bsnm{Ma}, \binits{C.}},
\oauthor{\bsnm{Yang}, \binits{H.H.}},
\oauthor{\bsnm{Farhad}, \binits{F.}},
\oauthor{\bsnm{Jin}, \binits{S.}},
\oauthor{\bsnm{Quek}, \binits{T.Q.S.}},
\oauthor{\bsnm{Poor}, \binits{H.V.}}:
Federated Learning with Differential Privacy: Algorithms and Performance Analysis.
{arXiv}.
\doiurl{10.48550/arXiv.1911.00222} .
\url{http://arxiv.org/abs/1911.00222}
Accessed 2024-04-12
\end{botherref}
\endbibitem

\bibitem[\protect\citeauthoryear{Agarwal et~al.}{}]{agarwal_skellam_2021}
\begin{botherref}
\oauthor{\bsnm{Agarwal}, \binits{N.}},
\oauthor{\bsnm{Kairouz}, \binits{P.}},
\oauthor{\bsnm{Liu}, \binits{Z.}}:
The skellam mechanism for differentially private federated learning.
In: Advances in Neural Information Processing Systems,
vol. 34,
pp. 5052--5064.
Curran Associates, Inc.
\url{https://papers.neurips.cc/paper/2021/hash/285baacbdf8fda1de94b19282acd23e2-Abstract.html}
Accessed 2023-12-03
\end{botherref}
\endbibitem

\bibitem[\protect\citeauthoryear{Mondal et~al.}{}]{mondal_beas_2022}
\begin{botherref}
\oauthor{\bsnm{Mondal}, \binits{A.}},
\oauthor{\bsnm{Virk}, \binits{H.}},
\oauthor{\bsnm{Gupta}, \binits{D.}}:
{BEAS}: Blockchain Enabled Asynchronous \& Secure Federated Machine Learning.
{arXiv}.
\url{http://arxiv.org/abs/2202.02817}
Accessed 2023-12-05
\end{botherref}
\endbibitem

\bibitem[\protect\citeauthoryear{Panda et~al.}{}]{panda_sparsefed_2021}
\begin{botherref}
\oauthor{\bsnm{Panda}, \binits{A.}},
\oauthor{\bsnm{Mahloujifar}, \binits{S.}},
\oauthor{\bsnm{Bhagoji}, \binits{A.N.}},
\oauthor{\bsnm{Chakraborty}, \binits{S.}},
\oauthor{\bsnm{Mittal}, \binits{P.}}:
{SparseFed}: Mitigating Model Poisoning Attacks in Federated Learning with Sparsification.
\url{http://arxiv.org/abs/2112.06274}
Accessed 2023-12-07
\end{botherref}
\endbibitem

\bibitem[\protect\citeauthoryear{Zhang et~al.}{}]{zhang_batchcrypt_nodate}
\begin{botherref}
\oauthor{\bsnm{Zhang}, \binits{C.}},
\oauthor{\bsnm{Li}, \binits{S.}},
\oauthor{\bsnm{Xia}, \binits{J.}},
\oauthor{\bsnm{Wang}, \binits{W.}},
\oauthor{\bsnm{Yan}, \binits{F.}},
\oauthor{\bsnm{Liu}, \binits{Y.}}:
{BatchCrypt}: Efﬁcient homomorphic encryption for cross-silo federated learning
\end{botherref}
\endbibitem

\bibitem[\protect\citeauthoryear{Mo et~al.}{}]{mo_ppfl_2021}
\begin{botherref}
\oauthor{\bsnm{Mo}, \binits{F.}},
\oauthor{\bsnm{Haddadi}, \binits{H.}},
\oauthor{\bsnm{Katevas}, \binits{K.}},
\oauthor{\bsnm{Marin}, \binits{E.}},
\oauthor{\bsnm{Perino}, \binits{D.}},
\oauthor{\bsnm{Kourtellis}, \binits{N.}}:
{PPFL}: Privacy-preserving Federated Learning with Trusted Execution Environments.
{arXiv}.
\url{http://arxiv.org/abs/2104.14380}
Accessed 2023-12-10
\end{botherref}
\endbibitem

\bibitem[\protect\citeauthoryear{Ioffe and Szegedy}{}]{ioffe_batch_2015}
\begin{botherref}
\oauthor{\bsnm{Ioffe}, \binits{S.}},
\oauthor{\bsnm{Szegedy}, \binits{C.}}:
Batch Normalization: Accelerating Deep Network Training by Reducing Internal Covariate Shift.
{arXiv}.
\url{http://arxiv.org/abs/1502.03167}
Accessed 2024-01-02
\end{botherref}
\endbibitem

\end{thebibliography}

\newpage

\begin{appendices}
\section{Full Algorithm of Obfuscated FL}\label{A}
\begin{algorithm}[th!]
    \caption{Obfuscated Federated Learning}
    \label{alg:FL_obfuscation}
    \begin{multicols}{2}
    \begin{algorithmic}

        \Require Central Server $S$, Central Model $M$, Client Models $C_1...C_n$, Client Data $D_1...D_n$, Aggregation Function $Agg(C*)$, \textbf{Obfuscation Method $Obf(C_i, p)$}, Number of Rounds $R$, Epochs per Round $E$
        \Ensure Each Client Data $D_i$ is only available to client $C_i$

        \State ...
                \For{$e$ in $1 \ldots E$}
                    \State $C_i \gets Fit(C_i, D_i)$ 
                \EndFor
                \State $C_i \gets Obf(C_i, p)$ 
                \State Send $C_i$ to $S$
        \State ...
        \State 
                
    \Procedure{Obf}{$C_i, p$}
    \If{$Obf \equiv Mask$ and $0 < p < 1$}
        \For{$i$ in $Parameters(C_i)$}
            \If{$rng(0,1) < p$}
                \State $i \gets NaN$ 
            \EndIf
        \EndFor
    \ElsIf{$Obf \equiv Noise$ and $0 < p$}
        \For{$i$ in $Parameters(C_i)$}
            \State $i \gets i + Gaussian(0, \sigma=p)$ 
        \EndFor
    \ElsIf{$Obf \equiv Clip$ and $0 < p < 1$}
        \For{$l$ in $Layers(C_i)$}
            \State $T \gets percentile(l, p)$
            \For{$i$ in $Parameters(l)$}
                \If{$|i| > T$}
                    \State $i \gets sign(i)*T$
                \EndIf
            \EndFor
        \EndFor
    \ElsIf{$Obf \equiv Prune$ and $0 < p < 1$}
        \For{$l$ in $Layers(C_i)$}
            \State $T \gets percentile(l, p)$
            \For{$i$ in $Parameters(l)$}
                \If{$|i| < T$}
                    \State $i \gets 0$
                \EndIf
            \EndFor
        \EndFor
        
    \EndIf
    \State \Return $Parameters(C_i)$
    \EndProcedure

    \end{algorithmic}
    \end{multicols}
\end{algorithm}
\end{appendices}

\end{document}